\documentclass[letterpaper]{article} 
\usepackage[preprint]{aaai2027}
\usepackage[hyphens]{url}  
\usepackage{graphicx} 
\usepackage{natbib}  
\usepackage{caption} 
\usepackage{amsmath,amsfonts,amssymb}
\usepackage{algorithm}
\usepackage{algorithmic}
\usepackage{array}
\usepackage{textcomp}
\usepackage{booktabs}
\usepackage{multirow}

\title{G2VD: Generalizable AI-Generated Video Detection via Counterfactual Intervention and Causal Disentanglement}
\author{
Meng Du,
Hongchang Chen,
Ran Li,
Junjie Zhang,\\
Qi Ouyang,
Shibo Zhang,
Shuxin Liu\corresponding
}
\affiliations{
Information Engineering University, Zhengzhou 450003, China\\
dm\_csy@alumni.nudt.edu.cn; liushuxin11@126.com
}

\begin{document}

\maketitle

\begin{abstract}
Rapid advances in AI video generation pose increasing security risks and call for reliable detectors with strong cross-domain generalization. Although existing methods perform well under in-domain evaluation, their performance degrades substantially on unseen generators. A key reason is shortcut learning, where detectors rely on domain-specific bias rather than intrinsic forensic cues. To address this issue, we propose G2VD, a generalizable AI-generated video detection framework based on counterfactual intervention and causal disentanglement. First, G2VD introduces a counterfactual intervention pipeline (CFIPipeline) that constructs counterfactual samples through VAE-based reconstruction and subsequent frequency-domain and pixel-domain alignment, thereby weakening spurious correlations between domain-specific bias and authenticity labels. Building on this intervention, we further design a causal disentanglement classifier that combines two domain-anchored branches with complementary objectives and a constraint based on the Hilbert-Schmidt Independence Criterion (HSIC), encouraging the causal and non-causal representations to capture intrinsic forensic cues and domain-specific bias, respectively. Experiments across four public datasets demonstrate strong cross-domain performance and consistent gains over baseline methods. In the challenging GenVidBench setting, G2VD achieves over 90\% overall ACC, with improvements of 0.194 in F1 and 0.104 in AUC over comparable state-of-the-art methods, while using only 10\% of the available training data. Code is available at \url{https://github.com/DMOSCAR-98/G2VD}.
\end{abstract}

\section{Introduction}

\begin{figure}[!t]
\centering
\includegraphics[width=0.95\columnwidth]{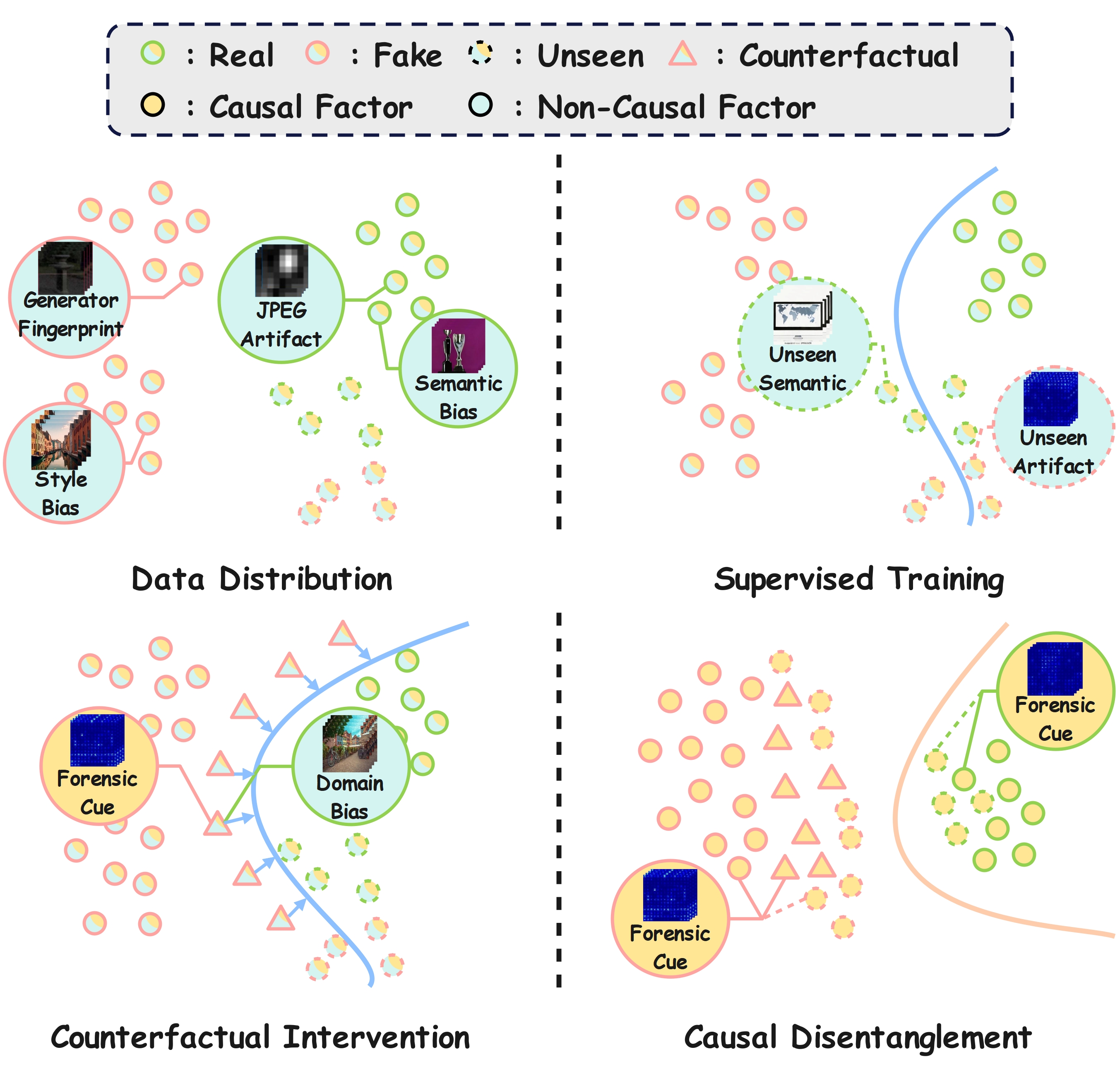}
\caption{Motivation of G2VD. Conventional training entangles intrinsic forensic cues with domain-specific bias, limiting generalization. Counterfactual intervention builds counterfactual samples that pair fake-video forensic cues with real-video domain characteristics, weakening spurious correlations between domain-specific bias and authenticity labels. Causal disentanglement separates causal from non-causal factors, and the former aids cross-domain generalization.}
\label{fig:motivation}
\end{figure}

Modern AI video generation models, such as Sora~\cite{brooks2024video}, Stable Video Diffusion~\cite{blattmann2023stable}, CogVideoX~\cite{yang2024cogvideox}, HunyuanVideo~\cite{kong2024hunyuanvideo}, and Wan~\cite{wan2025wan}, have made it increasingly easy to synthesize highly realistic videos from simple text or image inputs. While these technologies offer substantial creative potential, they also pose serious security risks, including the spread of misinformation, identity fraud, and manipulation of public opinion. Consequently, developing reliable AI-generated video detectors with strong cross-domain generalization has become increasingly urgent.

Early video forgery detectors primarily targeted facial deepfakes~\cite{tan2024frequency,wang2025fakediffer}. Subsequent work has extended detection to arbitrary-content videos, achieving promising performance by leveraging powerful video backbones and modeling spatiotemporal or higher-order inconsistencies~\cite{chen2024demamba,zheng2025d3}. More recently, multimodal large language model (MLLM)-based detectors have reframed authenticity verification as an explainable visual reasoning task, providing both authenticity predictions and supporting rationales~\cite{wen2025busterx,park2025vidguardr1}. However, large-scale benchmarks~\cite{ni2025genvidbench,ma2026aigvdbench,tang2026artifactbench} reveal persistent brittleness under generator-level distribution shifts, showing that MLLM-based methods may produce unreliable or hallucinated artifact explanations and do not consistently outperform specialized detectors. Accordingly, this work focuses on improving the cross-domain generalization of dedicated visual detectors.

A key source of this vulnerability is shortcut learning~\cite{geirhos2020shortcut}, whereby training on limited source domains encourages detectors to exploit domain-specific bias, such as generator fingerprints, compression patterns, and generation styles, rather than intrinsic forensic cues. As illustrated in Fig.~\ref{fig:motivation}, this entanglement yields decision boundaries that transfer poorly as spurious label correlations shift. Existing solutions use forensic-oriented augmentation~\cite{corvi2025seeing}, data alignment~\cite{chen2025dda}, and common-feature mining~\cite{kundu2025universal}, while causally guided feature disentanglement has been explored in the image domain~\cite{liu2026causalclip}. However, causally informed mechanisms for separating intrinsic forensic cues from domain-specific bias remain underexplored in AI-generated video detection, though such separation is key to cross-domain generalization.

To address this challenge, we develop G2VD, a counterfactual intervention and causal disentanglement framework for generalizable AI-generated video detection. Motivated by the structural causal model (SCM)~\cite{pearl2009causality}, G2VD treats intrinsic forensic cues as causal factors and domain-specific bias as non-causal factors. Accordingly, CFIPipeline combines VAE-based reconstruction with frequency-domain and pixel-domain alignment to construct counterfactual samples, thereby weakening spurious label correlations. A causal disentanglement classifier then uses two domain-anchored branches and an HSIC-based independence constraint~\cite{gretton2005measuring} to learn disentangled causal and non-causal representations from visual features, promoting transfer across diverse generators.

Our main contributions are summarized as follows:
\begin{itemize}
\item By rethinking AI-generated video detection from a causal perspective, we frame the entanglement between intrinsic forensic cues and domain-specific bias as a major source of cross-domain generalization failure.

\item We introduce G2VD, a generalizable AI-generated video detection framework that integrates reconstruction-based counterfactual intervention with domain-anchored causal disentanglement to mitigate shortcut learning and promote the learning of intrinsic forensic cues.

\item Extensive experiments on four public datasets demonstrate strong cross-domain performance of G2VD under a limited-data training protocol and validate the effectiveness of the proposed components.
\end{itemize}

\section{Related Work}

\subsection{AI-Generated Video Detection}

AI-generated video detection has progressed from face-oriented deepfake detectors that exploit frequency artifacts, reconstruction discrepancies, or spatiotemporal inconsistencies~\cite{tan2024frequency,wang2025fakediffer,yan2025generalizing} toward arbitrary-content detection for modern video generators. Dedicated detectors model frame consistency and spatiotemporal anomalies~\cite{ma2024gvf,bai2025gvd}, capture higher-order temporal features and long-range dependencies~\cite{zheng2025d3,chen2024demamba}, or learn common forensic cues from face or background manipulations and fully generated content~\cite{kundu2025universal}. Recent approaches further improve generalization through forensic-oriented frequency augmentation~\cite{corvi2025seeing}, native-scale artifact preservation~\cite{li2026preserving}, or cross-modal temporal modeling~\cite{wang2026cmta}. MLLM-based methods further frame authenticity verification as an explainable reasoning task and provide predictions with supporting explanations~\cite{wen2025busterx,park2025vidguardr1}. However, recent benchmarks~\cite{ni2025genvidbench,ma2026aigvdbench,tang2026artifactbench} show that under generator-level distribution shifts, existing methods still struggle to provide reliable authenticity judgments and artifact explanations.

\subsection{Causal Representation Learning}

Causal representation learning aims to identify stable predictive factors under distribution shifts~\cite{scholkopf2021toward}. SCM and do-calculus provide a foundation for intervention and counterfactual reasoning~\cite{pearl2009causality}. Related approaches learn invariant predictors~\cite{arjovsky2019invariant}, simulate interventions on factorized representations~\cite{lv2022cirl}, or apply causal disentanglement to deepfake detection~\cite{shi2026dynamic}. In generated-content forensics, CausalCLIP applies feature-level disentanglement and filtering to generated-image detection~\cite{liu2026causalclip}. Nevertheless, causally informed approaches remain underexplored in AI-generated video detection, particularly under generator-level domain shifts.

\subsection{Feature Disentanglement}

Feature disentanglement separates predictive signals from nuisance variations, typically using adversarial objectives~\cite{ganin2016domain}, mutual information regularization~\cite{chen2016infogan}, or independence criteria such as HSIC~\cite{gretton2005measuring}. In forgery detection, UCF separates manipulation-common from manipulation-specific components~\cite{yan2023ucf}, whereas multi-scale disentanglement based on critical forgetting suppresses source-specific forgery artifacts~\cite{li2025critical}. These methods primarily formulate feature decomposition in terms of common versus specific information or style versus content, rather than intrinsic forensic cues versus domain-specific bias under a causal formulation.

\section{Method}

\begin{figure*}[!t]
\centering
\includegraphics[width=0.95\textwidth]{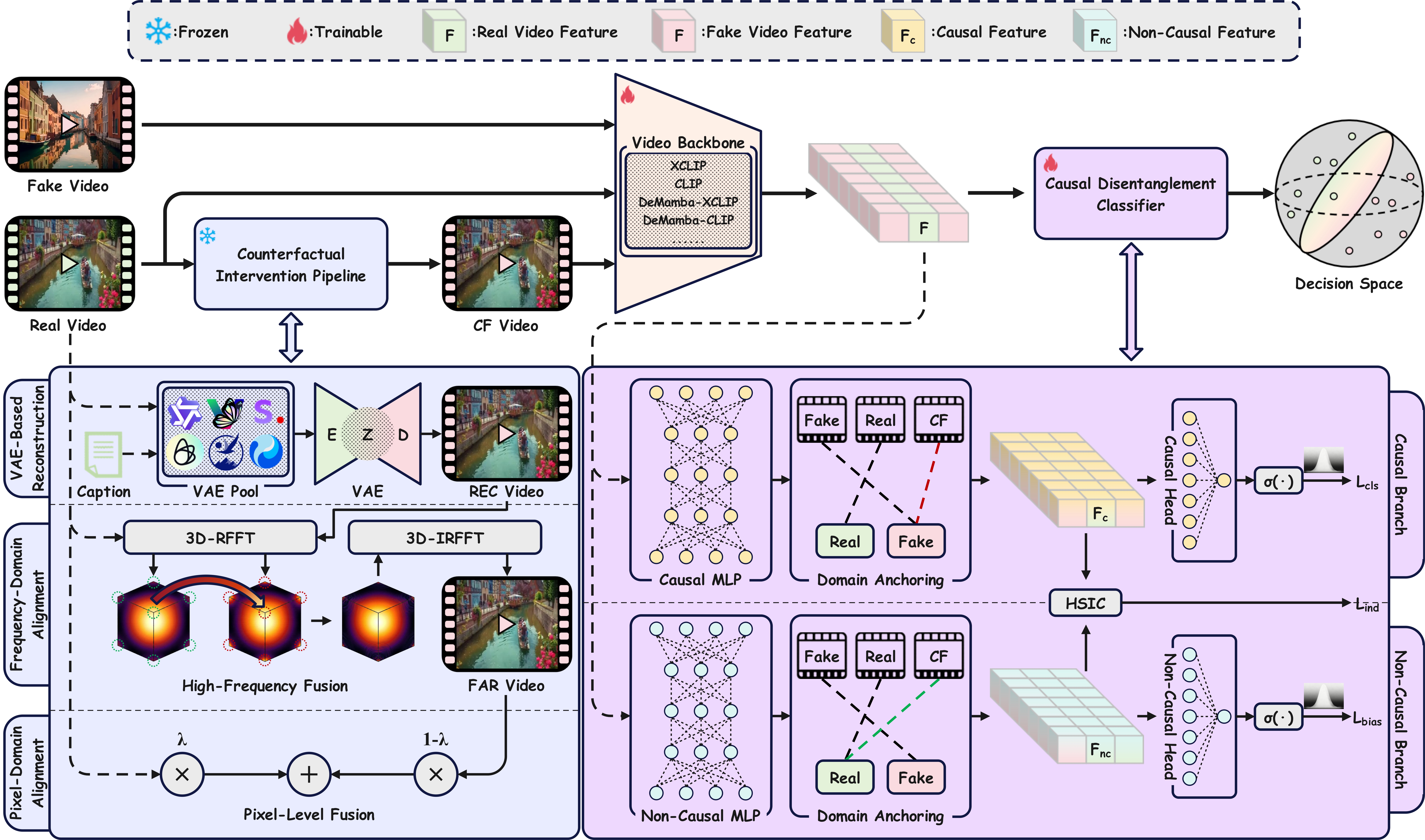}
\caption{Overview of G2VD. CFIPipeline constructs counterfactual videos through VAE-based reconstruction and frequency-domain and pixel-domain alignment. The video backbone extracts visual features $F$ from inputs. A causal disentanglement classifier then uses two complementary domain-anchored branches and an HSIC-based independence constraint to promote representation disentanglement, encouraging $F_c$ and $F_{nc}$ to capture intrinsic forensic cues and domain-specific bias, respectively. Inference uses only the backbone and causal branch. Snowflake and flame icons denote frozen and trainable components.}
\label{fig:framework}
\end{figure*}

As shown in Fig.~\ref{fig:framework}, G2VD integrates three modules: CFIPipeline constructs counterfactual samples; a video backbone extracts visual features $F$ from the input videos; and a causal disentanglement classifier maps $F$ to causal and non-causal representations. CFIPipeline remains frozen, while the backbone and classifier are jointly optimized.

\subsection{SCM-Based Causal Modeling}

Inspired by prior causality-based formulations for domain generalization~\cite{lv2022cirl}, we describe our task using the following SCM:
\begin{equation}
\left\{
\begin{aligned}
X &:= f(S,U,V_1), \quad S \perp\!\!\!\perp U \perp\!\!\!\perp V_1, \\
Y &:= h(S,V_2)=h(g(X),V_2), \quad V_1 \perp\!\!\!\perp V_2,
\end{aligned}
\right.
\end{equation}
where $X$ and $Y$ denote the video and label, respectively; $S$ denotes the causal factor associated with intrinsic forensic cues; $U$ denotes the non-causal factor associated with domain-specific bias, including semantic, source, and style characteristics; and $V_1,V_2$ are independent noise terms. The functions $f$, $g$, and $h$ represent video generation, feature extraction, and classification, respectively. Across distributions $P(X,Y)\in\mathcal{P}$, we assume that $P(Y\mid S)$ remains invariant, whereas correlations between $U$ and $Y$ may shift, motivating reliance on $S$ beyond the i.i.d.\ setting. With noise omitted, real and fake videos are represented as
\begin{equation}
\begin{cases}
X_r := f(S_r, U_r), & \text{Real}, \\
X_f := f(S_f, U_f), & \text{Fake}.
\end{cases}
\end{equation}
The domain-specific bias represented by $U_r$ and $U_f$ may support confident source-domain predictions, yet such reliance generalizes poorly as its label correlations shift. Accordingly, G2VD introduces the following CFIPipeline and causal disentanglement classifier to promote the separation of representations associated with $S$ and $U$.

\subsection{Counterfactual Intervention Pipeline}

Building on Eq.~(2), we consider an ideal counterfactual sample $X_{cf}^{*}$ that pairs the fake-video causal factor $S_f$ with the real-video non-causal factor $U_r$. Assigning $X_{cf}^{*}$ the fake label yields
\begin{equation}
\begin{cases}
X_r := f(S_r,U_r), & \text{Real}, \\
X_f := f(S_f,U_f), & \text{Fake}, \\
X_{cf}^{*} := f(S_f,U_r), & \text{Fake}.
\end{cases}
\label{eq:3}
\end{equation}
As $U_r$ is shared by $X_r$ and $X_{cf}^{*}$, it appears under both authenticity labels, weakening its correlation with authenticity. Meanwhile, the shared $S_f$ in $X_f$ and $X_{cf}^{*}$ encourages $F_c$ to capture intrinsic forensic cues across different domain-specific biases.

However, directly constructing $X_{cf}^{*}$ is challenging because $S_f$ and $U_r$ are latent and entangled in observed videos, making their direct isolation and recombination infeasible. To obtain an operational approximation, we exploit a common mechanism in modern video generation: VAEs encode videos into a latent space and decode them back, which can introduce reconstruction-induced traces. These traces provide a mechanistically motivated proxy signal associated with intrinsic forensic cues, without assuming that they are strictly equivalent to $S_f$. Thus, CFIPipeline first reconstructs $X_r$ through a VAE as
\begin{equation}
X_{rec} = \mathcal{D}(\mathcal{E}(X_r,t)),
\end{equation}
where $X_{rec}$ is the VAE-based reconstruction, $\mathcal{E}$ and $\mathcal{D}$ denote the VAE encoder and decoder, and $t$ is an optional text condition for conditional models. To discourage VAE-specific shortcut learning, we construct a pool of VAEs drawn from diverse video-generation frameworks and randomly sample one for each training batch.

On the other hand, motivated by previous work~\cite{chen2025dda}, we account for potential high-frequency attenuation in $X_r$ relative to the directly decoded $X_{rec}$ due to lossy video coding (e.g., H.264). We therefore use frequency-domain alignment to better preserve the domain characteristics associated with $U_r$ while retaining reconstruction-induced traces. RFFT is applied jointly over the temporal and spatial dimensions to capture video-level frequency structure:
\begin{equation}
\mathcal{F}_{r} = \mathrm{RFFT}(X_{r}), \quad
\mathcal{F}_{rec} = \mathrm{RFFT}(X_{rec}),
\end{equation}
where $\mathcal{F}_{r}$ and $\mathcal{F}_{rec}$ are spectra. Since phase carries spatiotemporal structure while compression mainly affects high-frequency magnitude, we align only amplitude:
\begin{equation}
A_{r} = |\mathcal{F}_{r}|, \quad
A_{rec} = |\mathcal{F}_{rec}|, \quad
\Phi_{rec} = \angle \mathcal{F}_{rec},
\end{equation}
where $A_r,A_{rec}$ are amplitudes and $\Phi_{rec}$ is the reconstructed phase. A high-pass mask selects the compression-sensitive region and fuses the amplitudes:
\begin{equation}
A_{fused} = M_{high} \odot A_{r} + (1 - M_{high}) \odot A_{rec}.
\end{equation}
$X_r$ supplies the high-frequency amplitudes, while $X_{rec}$ supplies the remaining spectral components. Recombining them through the inverse transform yields the frequency-aligned reconstruction $X_{far}$:
\begin{equation}
X_{far} = \mathrm{IRFFT}\left(A_{fused} \odot e^{i \Phi_{rec}}\right).
\end{equation}

Lastly, pixel-domain alignment uses interpolation to reduce residual color or contrast discrepancies and yield $X_{cf}$:
\begin{equation}
X_{cf} = \lambda X_{r} + (1-\lambda)X_{far},
\end{equation}
where $\lambda\in[0,0.5]$ controls the intervention strength. This final alignment reduces low-level appearance discrepancies while retaining reconstruction-induced traces, yielding $X_{cf}$ as an operational approximation to $X_{cf}^{*}$.

\subsection{Causal Disentanglement Classifier}

Counterfactual supervision alone may leave intrinsic forensic cues entangled with domain-specific bias. The causal disentanglement classifier addresses this by learning complementary representations from backbone features. Two domain-anchored branches assign distinct label semantics to features from $X_r$, $X_f$, and $X_{cf}$, while an HSIC-based independence constraint penalizes statistical dependence between the two branch representations.

The causal branch maps the backbone feature $F$ through a lightweight MLP to obtain the causal representation $F_c$:
\begin{equation}
F_c = \mathrm{MLP}_c(F).
\end{equation}

For domain anchoring in the causal branch, following Eq.~\eqref{eq:3}, $X_r$ is assigned $Y=0$, while $X_f$ and $X_{cf}$ are assigned $Y=1$. Thus, $X_f$ and $X_{cf}$ are assigned the same label despite their different domain characteristics. The branch is optimized by
\begin{equation}
\begin{split}
\mathcal{L}_{cls} = - \big[ & Y \log\big(\sigma(h_c(F_c))\big) \\
 & + (1-Y)\log\big(1-\sigma(h_c(F_c))\big) \big],
\end{split}
\end{equation}
where $h_c$ is the classification head and $\sigma$ denotes the sigmoid function. Minimizing $\mathcal{L}_{cls}$ discourages reliance on domain-specific bias and encourages $F_c$ to capture intrinsic forensic cues.

The non-causal branch uses a second MLP to obtain the non-causal representation $F_{nc}$:
\begin{equation}
F_{nc} = \mathrm{MLP}_{nc}(F).
\end{equation}

Its complementary domain-anchoring rule is
\begin{equation}
\begin{cases}
X_r := f(S_r,U_r), & \text{Real}, \\
X_f := f(S_f,U_f), & \text{Fake}, \\
X_{cf}^{*} := f(S_f,U_r), & \text{Real}.
\end{cases}
\end{equation}
Here $X_{cf}$ is relabeled real, contrary to the causal branch, so that samples sharing $U_r$ are anchored together despite different intrinsic forensic cues. The branch is optimized by
\begin{equation}
\begin{split}
\mathcal{L}_{bias} = - \big[ & Y' \log\big(\sigma(h_{nc}(F_{nc}))\big) \\
 & + (1-Y')\log\big(1-\sigma(h_{nc}(F_{nc}))\big) \big],
\end{split}
\end{equation}
where $h_{nc}$ is its classification head and $Y'$ follows the rule above. Minimizing $\mathcal{L}_{bias}$ encourages $F_{nc}$ to capture domain-specific bias. This complementary objective gives the two branches explicit semantic roles rather than relying on unconstrained feature partitioning.

\subsection{Objective Function}

The independence constraint is based on HSIC~\cite{gretton2005measuring}, a kernel measure of nonlinear dependence between mini-batch representations:
\begin{equation}
\mathcal{L}_{ind} = \mathrm{HSIC}(F_c, F_{nc}).
\end{equation}
Minimizing $\mathcal{L}_{ind}$ discourages overlapping information between the branches without requiring an adversarial discriminator. The overall objective is
\begin{equation}
\mathcal{L}_{total} = w_{cls} \mathcal{L}_{cls} + w_{bias} \mathcal{L}_{bias} + w_{ind} \mathcal{L}_{ind},
\end{equation}
where $w_{cls}$, $w_{bias}$, and $w_{ind}$ are loss weights. The first two terms encourage $F_c$ and $F_{nc}$ to capture intrinsic forensic cues and domain-specific bias, respectively, while $\mathcal{L}_{ind}$ discourages residual dependence not resolved by domain anchoring alone. Joint optimization promotes the separation of causal and non-causal representations for cross-domain detection. During inference, CFIPipeline and the non-causal branch are removed; prediction uses only the backbone and causal branch.

\section{Experiments}

In this section, we first describe the experimental protocol, including datasets, metrics, baselines, and implementation details; then evaluate cross-domain generalization against baseline methods; and finally analyze G2VD's components, branches, robustness, and feature distributions.

\providecommand{\tblbest}[1]{\textbf{#1}}
\providecommand{\tblsecond}[1]{\underline{#1}}
\providecommand{\extres}{\textsuperscript{*}}
\providecommand{\stat}[2]{}
\renewcommand{\stat}[2]{#1(#2)}

\begin{table*}[!t]
\centering
{\small
\setlength{\tabcolsep}{1.25pt}
\begin{tabular*}{0.998\textwidth}{@{\extracolsep{\fill}}ll*{5}{c}@{\hspace{5pt}\vrule width 0.4pt\hspace{-1pt}}cccc@{}}
\toprule
Method & Arch. & HD-VG & CogV & Mora & MuseV & SVD & \mbox{Ovr. ACC} & F1 & AUC & AP \\
\midrule
F3Net & CNN & \stat{88.6}{3.4} & \stat{45.5}{11.5} & \stat{84.4}{5.4} & \stat{24.1}{6.1} & \stat{19.7}{4.7} & \stat{52.7}{4.6} & \stat{.595}{.059} & \stat{.742}{.010} & \stat{.920}{.004} \\
STIL & CNN & \stat{74.5}{4.9} & \stat{51.8}{7.7} & \stat{86.4}{3.3} & \stat{33.9}{4.1} & \stat{29.8}{3.3} & \stat{55.5}{2.0} & \stat{.646}{.026} & \stat{.664}{.016} & \stat{.888}{.004} \\
\midrule
FTCN & TF & \stat{91.3}{2.6} & \stat{74.4}{1.1} & \stat{93.3}{0.6} & \stat{37.8}{3.2} & \stat{13.6}{0.8} & \stat{62.4}{0.7} & \stat{.702}{.008} & \stat{.802}{.011} & \stat{.945}{.004} \\
MINTIME & TF & \stat{93.7}{1.8} & \stat{80.9}{7.1} & \stat{92.0}{1.6} & \stat{29.8}{5.5} & \stat{10.8}{0.9} & \stat{61.8}{2.4} & \stat{.694}{.026} & \stat{.823}{.012} & \stat{.951}{.004} \\
TALL & TF & \stat{64.6}{3.0} & \stat{67.7}{12.2} & \stat{86.0}{1.1} & \stat{59.5}{3.8} & \stat{50.3}{1.3} & \stat{65.8}{2.3} & \stat{.756}{.022} & \stat{.710}{.020} & \stat{.900}{.007} \\
TimeSformer & TF & \stat{65.9}{3.8} & \stat{80.8}{2.2} & \stat{87.3}{1.4} & \stat{47.5}{7.2} & \stat{43.8}{5.5} & \stat{65.3}{2.5} & \stat{.751}{.025} & \stat{.712}{.008} & \stat{.908}{.002} \\
VideoMAE & TF & \stat{84.9}{4.5} & \stat{82.5}{4.1} & \stat{65.3}{9.9} & \stat{21.1}{7.4} & \stat{27.8}{9.2} & \stat{56.6}{3.7} & \stat{.646}{.045} & \stat{.733}{.003} & \stat{.920}{<.001} \\
ViViT & TF & \stat{65.2}{4.0} & \stat{76.9}{3.8} & \stat{90.4}{1.8} & \stat{43.9}{6.7} & \stat{46.2}{4.6} & \stat{64.8}{2.5} & \stat{.746}{.025} & \stat{.709}{.010} & \stat{.907}{.005} \\
\midrule
CLIP & TF & \stat{93.6}{1.7} & \stat{81.5}{5.2} & \stat{93.4}{1.9} & \stat{53.9}{12.3} & \stat{10.3}{2.9} & \stat{66.9}{3.9} & \stat{.744}{.039} & \stat{.845}{.009} & \stat{.959}{.002} \\
XCLIP & TF & \tblsecond{\stat{94.1}{1.1}} & \stat{82.9}{4.2} & \stat{93.2}{1.7} & \stat{32.5}{7.3} & \stat{9.3}{2.3} & \stat{62.8}{1.4} & \stat{.704}{.015} & \stat{.831}{.008} & \stat{.954}{.002} \\
DM-CLIP & Mamba & \stat{93.1}{1.0} & \stat{83.6}{4.0} & \stat{92.9}{0.7} & \stat{31.6}{2.2} & \stat{11.8}{0.5} & \stat{63.0}{0.9} & \stat{.707}{.009} & \stat{.810}{.011} & \stat{.949}{.003} \\
DM-XCLIP & Mamba & \tblbest{\stat{94.5}{1.0}} & \stat{83.6}{1.2} & \stat{91.3}{1.3} & \stat{25.9}{3.8} & \stat{9.7}{1.0} & \stat{61.4}{0.8} & \stat{.689}{.009} & \stat{.811}{.006} & \stat{.949}{.001} \\
\midrule
Qwen2.5-VL-7B\extres & MLLM & \stat{71.2}{--} & \stat{68.5}{--} & \stat{43.3}{--} & \stat{25.9}{--} & \stat{27.1}{--} & \stat{47.3}{--} & -- & -- & -- \\
GPT-4.1 mini\extres & MLLM & \stat{87.6}{--} & \stat{94.1}{--} & \stat{57.2}{--} & \stat{26.1}{--} & \stat{33.8}{--} & \stat{60.0}{--} & -- & -- & -- \\
VidGuard-R1\extres & MLLM & \stat{99.9}{--} & \stat{99.4}{--} & \stat{76.9}{--} & \stat{36.5}{--} & \stat{16.0}{--} & \stat{66.1}{--} & -- & -- & -- \\
\midrule
G2VD-CLIP & TF & \stat{79.9}{2.1} & \tblsecond{\stat{96.4}{2.3}} & \stat{99.7}{0.2} & \tblbest{\stat{93.0}{3.1}} & \tblbest{\stat{90.1}{2.8}} & \tblbest{\stat{91.9}{0.9}} & \tblbest{\stat{.950}{.006}} & \tblbest{\stat{.949}{.007}} & \tblbest{\stat{.982}{.002}} \\
G2VD-XCLIP & TF & \stat{81.1}{4.0} & \stat{94.8}{2.8} & \tblbest{\stat{99.9}{0.1}} & \stat{87.7}{3.4} & \stat{84.0}{8.3} & \stat{89.6}{1.4} & \stat{.934}{.010} & \stat{.940}{.010} & \stat{.981}{.004} \\
G2VD-DM-CLIP & Mamba & \stat{78.9}{2.4} & \tblbest{\stat{97.0}{2.5}} & \tblsecond{\stat{99.8}{0.1}} & \tblsecond{\stat{92.5}{2.5}} & \tblsecond{\stat{89.6}{2.4}} & \tblsecond{\stat{91.6}{0.9}} & \tblsecond{\stat{.948}{.006}} & \tblsecond{\stat{.947}{.008}} & \tblsecond{\stat{.981}{.003}} \\
G2VD-DM-XCLIP & Mamba & \stat{83.0}{3.9} & \stat{95.4}{2.3} & \stat{99.7}{0.2} & \stat{84.0}{3.7} & \stat{83.9}{3.4} & \stat{89.3}{0.5} & \stat{.932}{.004} & \stat{.927}{.008} & \stat{.973}{.004} \\
\bottomrule
\end{tabular*}
}
\caption{Cross-domain evaluation results on GenVidBench, reported as mean source-level ACC (\%), overall ACC (\%), F1, AUC, and AP over multiple seeds, with standard deviations in parentheses. * denotes external results derived from VidGuard-R1~\cite{park2025vidguardr1}. Bold and underlined values indicate the best and second-best results, respectively.}
\label{tab:genvidbench_results}
\end{table*}

\subsection{Experimental Setting}
\subsubsection{Datasets}
To evaluate cross-domain generalization, we use four public datasets: GenVidBench~\cite{ni2025genvidbench}, GenVideo~\cite{chen2024demamba}, GVD~\cite{bai2025gvd}, and GVF~\cite{ma2024gvf}. Our work uses the 143k version of GenVidBench, which contains two evaluation pairs. Pair1 comprises 20,131 real videos from Vript and 13,501 videos from each of four generators: Pika, VideoCrafter2, ModelScope, and T2V-Zero. Pair2 comprises 13,853 real videos from HD-VG and 13,853 videos from each of two text-to-video (T2V) models (Mora and CogVideo) and two image-to-video (I2V) models (MuseV and SVD). GenVideo is evaluated on its validation split of 8,588 videos from ten sources: Gen2, HotShot, LaVie, ModelScope, MoonValley, MorphStudio, Show1, Sora, VideoCrafter1, and the online-source WildScrape subset. GVD contains 11,618 videos from 11 sources across eight generators. Emu, HotShot, Sora, VideoCrafter1, and VideoPoet provide T2V sources; MoonValley, NeverEnds, and Pika each provide both T2V and I2V sources. GVF comprises 964 prompt-aligned real videos and generated counterparts from nine T2V generators: T2V-Zero, ModelScope, ZeroScope, Show1, Pika, Gen2, Sora, Veo, and Kling. Note that GenVidBench Pair2 is particularly challenging as its generated videos are conditioned on prompts or frames derived from corresponding real videos, resulting in high semantic and spatiotemporal similarity between video pairs, notably for MuseV and SVD.

\subsubsection{Evaluation Metrics}
We report dataset-level overall ACC, F1, AUC, and AP. Overall ACC denotes the Top-1 accuracy computed over all evaluated samples, avoiding distortions from equally averaging sources of unequal size. Source-level Top-1 ACC is also included to characterize performance on individual sources.

\subsubsection{Baselines}
The comparison covers four categories of detectors: CNN-based F3Net~\cite{qian2020thinking} and STIL~\cite{gu2021spatiotemporal}; transformer-based (TF) FTCN~\cite{zheng2021exploring}, MINTIME~\cite{coccomini2024mintime}, TALL~\cite{xu2023tall}, TimeSformer~\cite{bertasius2021timesformer}, ViViT~\cite{arnab2021vivit}, VideoMAE~\cite{tong2022videomae}, CLIP~\cite{radford2021learning}, and XCLIP~\cite{ni2022xclip}; Mamba-based DeMamba-CLIP and DeMamba-XCLIP~\cite{chen2024demamba}; and MLLM-based Qwen2.5-VL-7B, GPT-4.1 mini, and VidGuard-R1~\cite{park2025vidguardr1}. For compactness, DeMamba is abbreviated as DM.

\subsubsection{Implementation Details}
G2VD is instantiated with CLIP, XCLIP, DM-CLIP, and DM-XCLIP. The CFIPipeline pool contains 14 VAEs: 10 lightweight variants from the TAE projects~\cite{ollin2024taehv,ollin2024taesdv}, spanning frameworks such as HunyuanVideo~\cite{kong2024hunyuanvideo}, Wan~\cite{wan2025wan}, and LTX~\cite{hacohen2024ltxvideo}, together with 4 pretrained VideoVAE+ variants~\cite{xing2024videovaeplus}, some of which support text-conditioned reconstruction. For each video, we randomly sample a continuous 2-second clip, uniformly downsample it to 8 frames (16 for VideoMAE and 32 for ViViT), and resize to 224\texttimes224. All locally trained models are trained on a random 10\% subset of GenVidBench Pair1 and tested on a held-out subset. Although a few generators recur across datasets, the evaluated videos cover diverse content and generation settings beyond the sampled Pair1 training subset, supporting cross-domain evaluation. All experiments use PyTorch and four NVIDIA A800 GPUs (80GB) with four random seeds (40--43).

\providecommand{\tblbest}[1]{\textbf{#1}}
\providecommand{\tblsecond}[1]{\underline{#1}}
\providecommand{\extres}{\textsuperscript{*}}
\providecommand{\stat}[2]{}
\renewcommand{\stat}[2]{#1(#2)}

\begin{table}[!t]
\centering
{\small
\setlength{\tabcolsep}{1.5pt}
\begin{tabular*}{0.995\columnwidth}{@{\extracolsep{\fill}}llccc@{}}
\toprule
Method & Arch. & GenV & GVD & GVF \\
\midrule
F3Net & CNN & \stat{87.1}{3.3} & \stat{86.3}{3.9} & \stat{85.6}{3.5} \\
STIL & CNN & \stat{88.9}{1.8} & \stat{91.7}{0.9} & \stat{86.9}{2.1} \\
\midrule
FTCN & TF & \stat{93.5}{0.7} & \stat{94.5}{0.4} & \stat{95.7}{1.0} \\
MINTIME & TF & \stat{92.4}{0.8} & \stat{95.4}{1.0} & \stat{93.3}{1.3} \\
TALL & TF & \stat{93.9}{0.9} & \stat{96.0}{0.5} & \stat{93.5}{2.5} \\
TimeSformer & TF & \stat{95.7}{0.4} & \tblsecond{\stat{97.8}{0.4}} & \stat{91.1}{1.8} \\
VideoMAE & TF & \stat{83.3}{1.9} & \stat{87.6}{0.8} & \stat{91.4}{2.9} \\
ViViT & TF & \stat{94.8}{0.7} & \stat{95.7}{0.8} & \stat{93.2}{2.6} \\
\midrule
CLIP & TF & \stat{93.6}{0.9} & \stat{92.1}{1.9} & \stat{95.5}{1.8} \\
XCLIP & TF & \stat{93.3}{0.4} & \stat{94.9}{0.5} & \stat{93.8}{0.4} \\
DM-CLIP & Mamba & \stat{93.7}{0.8} & \stat{96.1}{0.8} & \stat{95.2}{0.5} \\
DM-XCLIP & Mamba & \stat{93.0}{0.5} & \stat{95.4}{0.8} & \stat{93.7}{1.2} \\
\midrule
Qwen2.5-VL-7B\extres & MLLM & \stat{54.3}{--} & -- & -- \\
GPT-4.1 mini\extres & MLLM & \stat{64.7}{--} & -- & -- \\
VidGuard-R1\extres & MLLM & \stat{89.9}{--} & -- & -- \\
\midrule
G2VD-CLIP & TF & \tblsecond{\stat{96.7}{0.5}} & \stat{95.0}{1.4} & \tblsecond{\stat{99.0}{0.3}} \\
G2VD-XCLIP & TF & \stat{96.4}{0.6} & \stat{96.8}{0.5} & \stat{98.3}{0.8} \\
G2VD-DM-CLIP & Mamba & \tblbest{\stat{97.5}{0.4}} & \tblbest{\stat{97.9}{0.2}} & \tblbest{\stat{99.4}{0.2}} \\
G2VD-DM-XCLIP & Mamba & \stat{96.7}{0.5} & \stat{97.2}{0.3} & \stat{98.4}{0.3} \\
\bottomrule
\end{tabular*}
}
\caption{Cross-domain evaluation results on GenVideo (GenV), GVD, and GVF, reported as mean overall ACC (\%) over multiple seeds. Symbols follow Table~\ref{tab:genvidbench_results}.}
\label{tab:genvideo_gvd_gvf_overall_results}
\end{table}

\subsection{Cross-Domain Evaluation on GenVidBench}
Table~\ref{tab:genvidbench_results} shows that G2VD-CLIP achieves the highest overall ACC of 91.9\%, with 0.950 F1 and 0.949 AUC; the other variants remain close at 89.3\%--91.6\%. Compared with the best baseline, G2VD-CLIP improves overall ACC by 25.0 pp. Each G2VD variant further improves its matched backbone by 25.0--28.6 pp, and different variants lead on different generated sources, indicating that the gain is not tied to one backbone's source preference. Improvements are most pronounced on the difficult MuseV and SVD sources, broadening source-level coverage. For the CLIP pair, this benefit is accompanied by a decrease on HD-VG from 93.6\% to 79.9\%, revealing a false-positive trade-off at the default operating point and motivating further calibration. Table~\ref{tab:genvidbench_results} compares VidGuard-R1's CoT variant; its potentially stronger GRPO variants are excluded from direct comparison due to their 7B parameter scale, different training data, and SFT/GRPO protocols, whereas G2VD variants use only 87--204M parameters and 10\% of GenVidBench Pair1 for training.

\subsection{Cross-Domain Evaluation on Other Datasets}
Table~\ref{tab:genvideo_gvd_gvf_overall_results} shows that G2VD-DM-CLIP ranks first on GenVideo, GVD, and GVF, reaching 97.5\%, 97.9\%, and 99.4\% overall ACC and exceeding the respective best baselines by 1.8, 0.1, and 3.7 pp. Every G2VD variant further improves its matched backbone by 1.8--4.7 pp across these datasets. The small absolute margin on GVD should be interpreted in the context of near-saturated baseline scores. Overall, the gains remain consistent across all four backbone pairs and three distinct generator mixtures. Together with the GenVidBench results, this consistency indicates a systematic improvement trend across architectures and datasets.

\begin{figure}[!t]
\centering
\includegraphics[width=0.90\columnwidth]{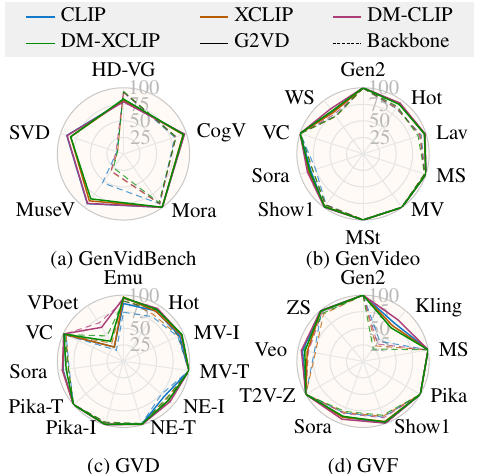}
\caption{Source-level evaluation results. Vertices denote generators; radii report mean ACC (\%) over multiple seeds.}
\label{fig:backbone_g2vd_radar}
\end{figure}

Figure~\ref{fig:backbone_g2vd_radar} complements the dataset-level results with source-level evidence. G2VD curves generally expand beyond their matched backbone curves, particularly on difficult GenVidBench, GenVideo, and GVF sources, indicating broader generator coverage rather than gains concentrated in a few easy sources. The agreement between dataset-level and source-level trends further shows that the aggregate improvement is not driven by a single dominant source.

\subsection{Ablation Studies}
To isolate each component, Table~\ref{tab:ablation_results} compares checkpoint-aligned variants: w/o CFI\&CD removes both CFIPipeline and the causal disentanglement classifier, w/o CD retains CFIPipeline only, and full G2VD includes both. Direct backbone fine-tuning performs substantially better on the other datasets than on GenVidBench, where w/o CFI\&CD achieves only 61.4\%--66.9\% overall ACC. This contrast highlights the difficulty of generalizing when real and generated videos share similar semantic and spatiotemporal content.
\providecommand{\tblbest}[1]{\textbf{#1}}
\providecommand{\tblsecond}[1]{\underline{#1}}
\providecommand{\stat}[2]{}
\renewcommand{\stat}[2]{#1(#2)}
\providecommand{\yes}{\ensuremath{\checkmark}}
\providecommand{\no}{\ensuremath{\times}}

\begin{table}[!t]
\centering
{\small
\setlength{\tabcolsep}{0pt}
\begin{tabular*}{0.995\columnwidth}{@{\extracolsep{\fill}}lcc*{4}{c}@{}}
\toprule
Backbone & CFI & CD & GVB & GenV & GVD & GVF \\
\midrule
\multirow{3}{*}{CLIP} & \no & \no & \stat{66.9}{3.9} & \stat{93.6}{0.9} & \stat{92.1}{1.9} & \stat{95.5}{1.8} \\
 & \yes & \no & \tblsecond{\stat{85.9}{2.6}} & \tblsecond{\stat{96.1}{0.7}} & \tblsecond{\stat{93.8}{2.2}} & \tblsecond{\stat{98.2}{0.6}} \\
 & \yes & \yes & \tblbest{\stat{91.9}{0.9}} & \tblbest{\stat{96.7}{0.5}} & \tblbest{\stat{95.0}{1.4}} & \tblbest{\stat{99.0}{0.3}} \\
\midrule
\multirow{3}{*}{XCLIP} & \no & \no & \stat{62.8}{1.4} & \stat{93.3}{0.4} & \stat{94.9}{0.5} & \stat{93.8}{0.4} \\
 & \yes & \no & \tblsecond{\stat{85.7}{1.0}} & \tblsecond{\stat{96.2}{0.8}} & \tblbest{\stat{97.0}{1.3}} & \tblsecond{\stat{97.8}{1.0}} \\
 & \yes & \yes & \tblbest{\stat{89.6}{1.4}} & \tblbest{\stat{96.4}{0.6}} & \tblsecond{\stat{96.8}{0.5}} & \tblbest{\stat{98.3}{0.8}} \\
\midrule
\multirow{3}{*}{DM-CLIP} & \no & \no & \stat{63.0}{0.9} & \stat{93.7}{0.8} & \stat{96.1}{0.8} & \stat{95.2}{0.5} \\
 & \yes & \no & \tblsecond{\stat{85.9}{0.7}} & \tblsecond{\stat{97.4}{0.6}} & \tblsecond{\stat{97.6}{0.5}} & \tblsecond{\stat{99.0}{0.2}} \\
 & \yes & \yes & \tblbest{\stat{91.6}{0.9}} & \tblbest{\stat{97.5}{0.4}} & \tblbest{\stat{97.9}{0.2}} & \tblbest{\stat{99.4}{0.2}} \\
\midrule
\multirow{3}{*}{DM-XCLIP} & \no & \no & \stat{61.4}{0.8} & \stat{93.0}{0.5} & \stat{95.4}{0.8} & \stat{93.7}{1.2} \\
 & \yes & \no & \tblsecond{\stat{83.2}{2.8}} & \tblsecond{\stat{95.9}{0.9}} & \tblsecond{\stat{96.3}{1.8}} & \tblsecond{\stat{97.3}{1.0}} \\
 & \yes & \yes & \tblbest{\stat{89.3}{0.5}} & \tblbest{\stat{96.7}{0.5}} & \tblbest{\stat{97.2}{0.3}} & \tblbest{\stat{98.4}{0.3}} \\
\bottomrule
\end{tabular*}
}
\caption{G2VD ablation results on GenVidBench (GVB), GenVideo (GenV), GVD, and GVF, reported as mean overall ACC (\%) over multiple seeds. $\checkmark$ and $\times$ indicate active and disabled components. Symbols follow Table~\ref{tab:genvidbench_results}.}
\label{tab:ablation_results}
\end{table}

CFIPipeline provides the dominant improvement, raising average overall ACC by 21.7 pp on GenVidBench and 2.7 pp across the other datasets. Adding CD yields further average gains of 5.4 and 0.6 pp. The larger improvements on GenVidBench, where real and generated videos are closely aligned in content, are consistent with intervention and disentanglement reducing reliance on domain-specific bias. Their consistent direction across all backbones supports complementary roles for the two components.

\subsection{Experimental Analysis}
\begin{figure}[!t]
\centering
\includegraphics[width=0.94\columnwidth]{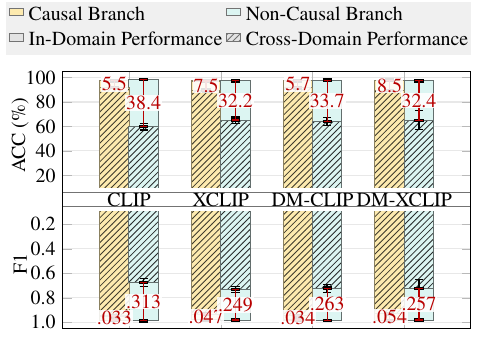}
\caption{Branch-wise in-domain and cross-domain performance on GenVidBench, reported as overall ACC (\%) and F1. Error bars denote mean and standard deviation over multiple seeds; red endpoint lines indicate the performance gap.}
\label{fig:branch_generalization}
\end{figure}

\begin{figure}[!t]
\centering
\includegraphics{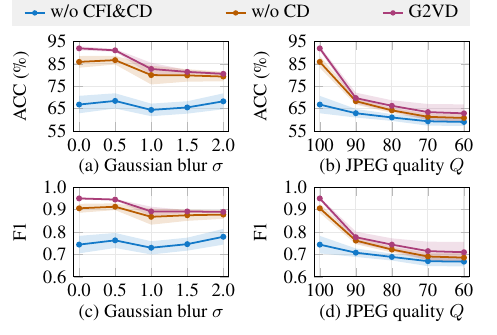}
\caption{Robustness evaluation results of CLIP-based variants on GenVidBench under Gaussian blur $\sigma$ and JPEG quality $Q$, reported as overall ACC (\%) and F1. Curves and bands denote mean and standard deviation over multiple seeds.}
\label{fig:robustness_lines}
\end{figure}

\begin{figure}[!t]
\centering
\includegraphics{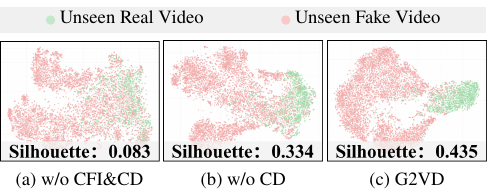}
\caption{t-SNE of CLIP-based variants on GenVidBench using seed 42. Silhouette scores are shown in the panels.}
\label{fig:feature_distribution}
\end{figure}

\subsubsection{Branch Generalization Analysis}
Figure~\ref{fig:branch_generalization} compares the causal and non-causal branches on Pair1 (in-domain) and Pair2 (cross-domain). Both perform similarly on Pair1, but on Pair2 the causal branch retains around 90\% ACC and above 0.93 F1, with ACC and F1 gaps of only 5.5--8.5 pp and 0.033--0.054, compared with 32.2--38.4 pp and 0.249--0.313 for the non-causal branch. This consistent pattern across backbones suggests that CD guides the causal branch toward intrinsic forensic cues and the non-causal branch toward domain-specific bias, supporting the intended disentanglement.
\subsubsection{Robustness Evaluation}
Figure~\ref{fig:robustness_lines} shows gradual degradation under Gaussian blur but a sharper decline under JPEG compression, while G2VD remains strongest throughout. At $Q=60$, its ACC and F1 fall from 91.9\% and 0.950 to 62.9\% and 0.710. This sensitivity suggests that aggressive compression removes part of the forensic evidence used by the detector and motivates post-processing-aware training.
\subsubsection{Feature Distribution Visualization}
Using seed 42, Fig.~\ref{fig:feature_distribution} shows progressively clearer real--fake separation from the backbone-only variant through CFIPipeline to full G2VD. To quantify this progression more reliably, we compute the silhouette score~\cite{rousseeuw1987silhouettes} in the high-dimensional feature space before t-SNE; it rises from 0.083 to 0.334 and 0.435. This component-wise progression complements the ablation and branch analyses, showing that the component gains coincide with increasingly separable representations.

The supplementary material shows detailed dataset statistics and more source-level, robustness, and t-SNE results.

\section{Conclusion}
In this paper, we introduce G2VD, a generalizable AI-generated video detection framework via counterfactual intervention and causal disentanglement. CFIPipeline weakens spurious correlations between domain-specific bias and authenticity labels, while the causal disentanglement classifier promotes the separation of causal and non-causal representations. Experiments across four datasets show consistent gains over baseline methods under a limited-data protocol, and ablations support both components. These results demonstrate that counterfactual intervention and causal disentanglement can guide detectors away from generator-specific shortcuts and toward intrinsic forensic cues, providing a practical basis for cross-domain video forensics. Future work will explore real-world deployment by further improving model robustness and calibration.

\section*{Acknowledgments}
This work was supported by the Top Talent Cultivation Program of Henan Province under Grant 244500510012.

\bibliography{references}

\end{document}